\documentclass[journal]{IEEEtran}
\usepackage{amsmath,amsfonts}
\usepackage{multirow}
\usepackage{algorithm}
\usepackage[noend]{algpseudocode}
\usepackage{array}
\usepackage[caption=false,font=normalsize,labelfont=sf,textfont=sf]{subfig}
\usepackage{textcomp}
\usepackage{stfloats}
\usepackage{verbatim}
\usepackage{graphicx}
\usepackage{cite}
\usepackage[table]{xcolor}
\usepackage{adjustbox}

\usepackage{hyperref}
\hypersetup{
	colorlinks=true,
	linkcolor=blue,
	filecolor=magenta,      
	urlcolor=blue,
	pdftitle={Overleaf Example},
	pdfpagemode=FullScreen,
}

\def\BibTeX{{\rm B\kern-.05em{\sc i\kern-.025em b}\kern-.08em
		T\kern-.1667em\lower.7ex\hbox{E}\kern-.125emX}}

\begin{document}
\title{Greedy Shapley Client Selection for Communication-Efficient Federated Learning}

\markboth{IEEE Journal Submission}%
{}
\author{Pranava Singhal, %
	\thanks{Pranava Singhal is with the Electrical Engineering Department, IIT Bombay, Mumbai, India Email: \{pranava.psinghal\}@gmail.com.} Shashi Raj Pandey,%
	\thanks{Shashi Raj Pandey and Petar Popovski are with the Connectivity Section, Department of Electronic Systems, Aalborg University, Denmark. Email: \{srp, petarp\}@es.aau.dk.}
	\textit{IEEE Member}, %
	and Petar Popovski, \textit{IEEE Fellow} 
	\thanks{This work was supported in parts by the Villum Investigator Grant “WATER” from the Velux Foundation, Denmark, and the European Union’s Horizon project 6G-XCEL (101139194).}
}

\maketitle

\begin{abstract}
	
	The standard client selection algorithms for Federated Learning (FL) are often unbiased and involve uniform random sampling of clients. This has been proven sub-optimal for fast convergence under practical settings characterized by significant heterogeneity in data distribution, computing, and communication resources across clients. For applications having timing constraints due to limited communication opportunities with the parameter server (PS), the client selection strategy is critical to complete model training within the fixed budget of communication rounds. To address this, we develop a biased client selection strategy, \textsc{GreedyFed}, that identifies and greedily selects the most contributing clients in each communication round. This method builds on a fast approximation algorithm for the Shapley Value at the PS, making the computation tractable for real-world applications with many clients. Compared to various client selection strategies on several real-world datasets, \textsc{GreedyFed} demonstrates fast and stable convergence with high accuracy under timing constraints and when imposing a higher degree of heterogeneity in data distribution, systems constraints, and privacy requirements.
\end{abstract}


\section{Introduction}
\IEEEPARstart{F}{ederated} Learning (FL) \cite{mcmahan2017communication} is a paradigm for distributed training of neural networks where a resourceful parameter server (PS) aims to train a model using the privately held data of several distributed clients. This is done through iterative interaction between the PS and clients providing the output of their local compute, i.e., local models, instead of raw data; hence, FL is often described as privacy-preserving model training. In such an approach, the communication bottleneck constrains the PS to (1) communicate with a few clients and (2) complete training in the fewest possible communication rounds. Our work focuses on developing a client selection strategy by which the PS can minimize communication overhead and speed up model convergence. 

The conventional scheme for FL, \textsc{FedAvg} \cite{mcmahan2017communication}, involves uniform random sampling of a subset of clients in each communication round to perform local computing over their private data. Upon receiving the local models from the selected clients, the PS then aggregates the updates by uniformly averaging model weights (or in proportion to local training data quantity) and initiates the next communication round. \textsc{FedAvg} proposes to reduce the total number of communication rounds through multiple epochs of client-side training in each round before model averaging. However, averaging model updates from uniformly sampled clients may lead to poor convergence of the PS model in typical FL settings with a high degree of heterogeneity in client data distribution, computational resources, and communication bandwidth.

To overcome these challenges with \textsc{FedAvg} several biased client selection strategies have been proposed for FL, focusing on improving the convergence rate and reducing total communication rounds while ensuring robustness to heterogeneity.

\subsection{Prior Work}
To minimize the communication overhead, existing works focus on either of the two approaches: (1) model compression through techniques such as quantization, pruning, or distillation \cite{konevcny2016federated, stich2018sparsified, cheng2017survey}, where client updates are compressed to reduce the bandwidth usage in each round of communication with the PS, and (2) reduction of the total number of communication rounds to achieve satisfactory accuracy (or convergence). This is done through modification of the local update scheme, such as in \textsc{FedProx} \cite{li2020federated} and \textsc{SCAFFOLD} \cite{karimireddy2020scaffold}, or using biased client selection strategies, such as  \textsc{Power-Of-Choice} \cite{cho2020client} and \textsc{IS-FedAvg} \cite{rizk2021optimal}. Algorithms like \textsc{FedProx} introduce a squared penalty term in the client loss to improve personalization in a communication-efficient manner, and \textsc{SCAFFOLD} uses control variates to prevent the client models from drifting away from the PS model. Most modifications to develop an efficient client selection strategy allow performing local updates in the same way as \textsc{FedAvg}. For instance, \textsc{Power-Of-Choice} selects clients with the highest local loss after querying all models for loss in each round, and \textsc{IS-FedAvg} modifies client selection and local data subset selection probabilities to minimize gradient noise. Our algorithm \textsc{GreedyFed} focuses on developing a novel biased client selection strategy adopting game-theoretic data valuation principles to improve communication efficiency in FL.

Many prior works have modeled client participation in FL as a cooperative game and looked at the Shapley-Value (SV) \cite{shapley1971cores} as a measure of client contribution to training. SV is an attractive choice due to its unique properties of \emph{fairness}, \emph{symmetry}, and \emph{additivity}. While related works have used SV for client valuation and incentivizing participation \cite{wang2019measure, wang2020principled} for collaborative training, a few works have also used SV for client selection \cite{pandey2023goal, nagalapatti2021game, huang2021shapley} to improve robustness and speed up training. Two key works using SV for client selection are: \cite{pandey2023goal}, in which the authors propose an Upper-Confidence Bound (UCB) client selection algorithm, and \cite{nagalapatti2021game}, which biases the client selection probabilities as softmax of Shapley values. A key limitation of these works is a lack of performance analysis of SV-based client selection in the presence of timing constraints and heterogeneity in data, systems, and privacy requirements, which are essential considerations for the practical deployment of FL algorithms. 

\subsection{Main Contributions}
In this paper, we develop a novel greedy SV-based client selection strategy for communication-efficient FL, called \textsc{GreedyFed}. It requires fewer communication rounds for convergence than state-of-the-art techniques in FL, including SV-based strategies, by involving the most contributing clients in training. Our algorithm builds on \cite{pandey2023goal} and \cite{nagalapatti2021game} with two essential modifications to enhance training efficiency:\\
(1) Purely greedy client selection (motivated in Section~\ref{ss:motivation})\\
(2) Integration of a fast Monte Carlo Shapley Value approximation algorithm \textsc{GTG-Shapley} \cite{liu2022gtg} making SV computation tractable for practical applications with several clients.

We demonstrate that \textsc{GreedyFed} alleviates the influence of constrained communication opportunities between clients and the PS, addressing practical deployment challenges due to premature completion of model training and its impact on model accuracy, and is robust towards heterogeneity in data, system constraints, and privacy requirements.  

\emph{Organization:} The rest of the paper is organized as follows. In Section~\ref{ss:setup}, we describe the client selection problem and set up notation. Section~\ref{ss:algorithm} outlines the details of our \textsc{GreedyFed} algorithm. In section~\ref{ss:experiments}, we describe the experiments, and in Section~\ref{ss:analysis} we analyze the results. Section~\ref{ss:conclusion} concludes the paper with directions for future work.

\section{Problem Setup}\label{ss:setup}
Consider a resourceful parameter server (PS) collaboratively training a machine learning model leveraging the local data and compute of $N$ resource-constrained clients. Each client $k \in [N] = \{1,\, 2,\, \cdots,\, N\}$ holds a private dataset $\mathcal{D}_k$ with $n_k$ samples. 
Our goal is to design a client selection strategy that selects a subset of $M$ clients, denoted $S_t$, in each communication round $t$ to train the PS model $w$ that gives the best prediction $f(\boldsymbol{x}, w)$ while generalizing well on a test set $\mathcal{D}_{\textrm{test}}$ in the fewest communication rounds. Let $\mathcal{D}_{\textrm{train}}$ denote the union of all client datapoints $\bigcup_{k \in [N]} \mathcal{D}_k$ with total training samples $n_{\textrm{train}} = \sum_{k \in [N]}n_k = |\mathcal{D}_{\textrm{train}}|$. The model $w$ is then obtained such as it minimizes the overall training loss, defined as
\begin{align*}
	\text{Training loss: }\mathcal{L}(w; \mathcal{D}_{\textrm{train}}) = \frac{1}{n_{\textrm{train}}}\sum_{(\boldsymbol{x},y) \in \mathcal{D}_{\textrm{train}}}\ell(y, f(\boldsymbol{x}, w)) \\
	= \sum_{k \in [N]} \frac{n_k}{n_{\textrm{train}}} \sum_{(\boldsymbol{x},y) \in \mathcal{D}_{k}}\frac{1}{n_k}\ell(y, f(\boldsymbol{x}, w)) = \sum_{k \in [N]} q_k \mathcal{L}(w; \mathcal{D}_k),
\end{align*}
where $\ell(y, f(\boldsymbol{x}, w))$ is the loss for a single datapoint. In round $t$, the server aggregates the updated model parameters $w_k^{(t+1)}$ from selected clients $k \in S_t$ and obtains the model $w^{(t+1)} = \sum_{k \in S_t} \lambda_k w_k^{(t+1)}$, with weights $\lambda_k$ proportional to the size of the client's dataset $n_k$ and summing to one. 

In our method, the server utilizes the SV to quantify selected clients' average contribution for selection in further rounds. Our method operates in two stages: initial client valuation and greedy client selection (c.f. Section~\ref{ss:cs_strategy}, \ref{ss:motivation}). 

In the following, we first discuss data valuation principles to quantify client valuation.

\emph{Shapley-Value for Client Valuation:} We model FL as an $N$ player cooperative game. When a subset $S_t$ of $M < N$ clients are selected in round $t$, Shapley-Value \cite{shapley1971cores} measures each player's average marginal contribution over all subsets $S$ of $S_t$. $U$ denotes a utility function on $2^{[M]} \to \mathbb{R}$, which associates a reward/value with every subset of clients. In FL, a natural choice for the utility function is negative of validation loss, which can be evaluated at the server from aggregated client updates $\{w_k^{(t+1)}\}_{k \in S_t}$ and a validation dataset $\mathcal{D}_{\textrm{val}}$ held at the server. This choice aligns with our objective of minimizing model loss. Thus, for a subset of clients $S$, utility $U(S)$ is computed on the weighted average of client model updates in $S$. Then, the Shapley Value of client $k$ is defined as follows:
\begin{align*}
	SV_k = \frac{1}{M}\sum_{S \in [M]\setminus k} \frac{U(S \cup k) - U(S)}{\binom{M-1}{|S|}}.
\end{align*}

Due to the combinatorial nature of SV computation, it becomes infeasible for many clients, and several Monte Carlo (MC) sampling approximations have been proposed \cite{liu2022gtg}\cite{ghorbani2019data}. \cite{liu2022gtg} shows that Monte Carlo SV approximation in a single round suffers from computational complexity of $\mathcal{O}(M \log M)$. \textsc{GTG-Shapley} reduces complexity by truncating MC sampling, achieving upto $\mathcal{O}(\log M)$ in the case of IID clients, and outperforms other SV approximation schemes in practice.

Client contributions in round $t$, denoted $SV_{k}^{(t)}$, are used to update the value of clients $k \in S_t$. Since practical FL settings only have partial client participation (selecting $M$ out of $N$ clients), we need to compute a cumulative Shapley-Value to compare all clients. \cite{wang2020principled} proposes to compute the cumulative Shapley Value as $SV_k = \sum_{t = 1}^{T} SV_{k}^{(t)}$ with $SV_{k}^{(t)} = 0$ for rounds where client $k$ is not selected. However, this will unfavourably value clients that were selected less often. Client selection algorithms \textsc{S-FedAvg}\cite{nagalapatti2021game} and \textsc{UCB}\cite{pandey2023goal} use the mean of $SV_{k}^{(t)}$ over $t$ where client $k$ was selected, to define cumulative SV, and we borrow this choice for our algorithm. We introduce a greedy selection strategy using cumulative SV incorporating a fast approximation algorithm \textsc{GTG-Shapley} \cite{liu2022gtg} to make SV computation tractable. 

\section{Proposed Method}\label{ss:algorithm}
This section discusses the proposed client selection algorithm for FL, \textsc{GreedyFed}, which greedily selects clients based on cumulative SV. 
\subsection{Client Selection Strategy}\label{ss:cs_strategy}
\textsc{GreedyFed} greedily selects the $M$ clients with the largest cumulative Shapley-Value $SV^{(t)}$ in each communication round $t$. Our selection scheme has no explicit exploration in the selection process. In contrast, the previous approach \textsc{UCB} \cite{pandey2023goal} proposed an Upper-Confidence Bound client selection criterion with an explicit exploration term added to the cumulative SV. In \textsc{S-FedAvg} \cite{nagalapatti2021game}, client selection probabilities are assigned as a softmax over a value vector. This value vector is computed as an exponentially weighted average of the cumulative Shapley Values. The softmax sampling in \textsc{S-FedAvg} allows for exploration by sometimes sampling clients with lower past contributions with non-zero probability. 

Our client selection procedure undergoes two operations: (1) Initialise Shapley Values using round-robin sampling for $M$ clients per round until $SV_k$ is initialised for all clients $k \in [N]$, and (2) subsequently perform greedy selection of top $M$ clients based on $SV_k$. We consider two variants for cumulative SV with exponentially weighted and standard averaging. 

Round-robin sampling allows initial valuation of all clients, and subsequent greedy selection allows for rapid convergence, minimising communication overhead. Next, we discuss the exploration-exploitation tradeoff motivating our design.

\subsection{Exploration-Exploitation Tradeoff}\label{ss:motivation}
This design is motivated by two reasons: \emph{First}, Round-robin (RR) sampling ensures that every client is selected at least once before comparing client qualities. Exploiting early leads to longer discovery times for unseen high-quality clients, especially without prior knowledge of the client reward distribution. This is experimentally observed in the case of \textsc{S-FedAvg}, which starts exploiting early but is rapidly superseded by \textsc{UCB} and \textsc{GreedyFed} after RR sampling is over. \emph{Second}, a client's valuation in FL decreases as it is selected more often. $SV_k^{(t)}$ is bounded by the decrease in validation loss in round $t$ due to the additivity property of SV that gives $\sum_{k \in S_t}{SV_{k}^{(t)}} = \mathcal{L}(w^{(t)}; \mathcal{D}_{\textrm{val}}) - \mathcal{L}(w^{(t+1)}; \mathcal{D}_{\textrm{val}})$. Thus, as the model converges, the validation loss gradient approaches zero, and thus $SV_{k}^{(t)}$ obeys a decreasing trend in $t$, and the clients selected frequently and recently experience a decrease in average value $SV_k$. This leads \textsc{GreedyFed} to fall back to other valuable clients after exploiting the top-$M$ clients and not the less explored clients like \textsc{UCB}. This explains the experimentally observed performance improvement of \textsc{GreedyFed} over \textsc{UCB}, despite identical RR initialisation.

The client selection and fast SV computation algorithms are outlined in Alg.~\ref{alg:cap} and Alg.~\ref{alg:cu}. The \textit{ModelAverage}$(n_k, w_k)$ subroutine computes a weighted average of models $w_k$ with weights proportional to $n_k$, and \textit{ClientUpdate} performs training on client data $\mathcal{D}_k$ starting from the current server model.

\begin{figure}[h!]
	\begin{algorithm}[H]
		\caption{Greedy Shapley-based Client Selection $\text {GreedyFed}(\{\mathcal{D}_k\},\, \mathcal{D}_{\textrm{val}},\,w^{(0)}, \, T)$}\label{alg:cap}
		\textbf{Input}: $N$ clients with datasets $\{\mathcal{D}_k\}_{k = 1}^{N}$, server with validation dataset $\mathcal{D}_{\textrm{val}}$, initial model weight $w^{(0)}$, number of communication rounds $T$, client selection budget $M$\\
		\textbf{Hyperparameters}: Training epochs per round $E$, Mini-batches per training epoch $B$, learning rate $\eta$, momentum $\gamma$, exponential averaging parameter $\alpha$\\
		\textbf{Output}: Trained model $w^{(T)}$\\
		\textbf{Initialise}: Client selections $N_k = 0,\, \forall\, k \in [N]$
		\begin{algorithmic}[1]
			\For {$t = 0,\, 1,\, 2,\, \cdots,\, T-1$}
			\If {$t < \lceil \frac{N}{M} \rceil$} Round-Robin (random order)
			\State $S_t = \{t,\, t+1,\, \cdots,\, t+M-1\}$
			\Else { Greedy Selection}
			\State $S_t = M \text{ largest SV clients}$
			\EndIf
			
			\For{client $k$ in $S_t$}
			\State $w_{k}^{(t+1)} = $ ClientUpdate$(\mathcal{D}_k,\, w^{(t)};\,E,\, B,\, \eta,\, \gamma)$
			\State $N_k \gets N_k + 1$
			\EndFor
			\State $w^{(t+1)} = $ ModelAverage$(n_k,\, w_k^{(t+1)} : k \in S_t)$
			\State $\{SV_k^{(t)}\}_{k \in S_t} = $ GTG-Shapley$(w^{(t)},\,\{w_{k}^{(t+1)}\},\, \mathcal{D}_{\textrm{val}})$
			\For{client $k$ in $S_t$}
			\State $SV_k \gets \frac{(N_k - 1)SV_k\, +\, SV_k^{(t)}}{N_k}$ (Mean) OR
			\Statex \hspace*{\algorithmicindent}\hspace*{\algorithmicindent}$SV_k \gets \alpha \cdot SV_k\, +\, (1 - \alpha)\cdot SV_k^{(t)}$ (Exponential)
			\EndFor
			\EndFor \\
			\textbf{return} $w^{(T)}$
		\end{algorithmic}
	\end{algorithm}
	\label{fig:greedyfedalgo}
	\vspace{-2\baselineskip}
\end{figure}

\begin{figure}[h!]
	\begin{algorithm}[H]
		\caption{Server-Side Shapley Value Approximation $\text {GTG-Shapley}(w^{(t)},\,\{w_{k}^{(t+1)}\}_{k \in S_t},\, \mathcal{D}_{\textrm{val}})$}\label{alg:cu}
		\textbf{Input}: Current server model $w^{(t)}$, client updated models $\{w_{k}^{(t+1)}\}$, validation dataset at server $\mathcal{D}_{\textrm{val}}$\\
		\textbf{Hyperparameters} : Error threshold $\epsilon$, maximum iterations $T$\\
		\textbf{Output} : Shapley Values $\{SV_k\}_{k \in S_t}$\\
		\textbf{Initialise} : $SV_k = 0, \, \forall \, k \in S_t$
		\begin{algorithmic}[1]
			\State Compute $w^{(t+1)} = $ ModelAverage$(n_k,\, w_k^{(t+1)} : k \in S_t)$
			\State $v_0 = \mathcal{U}(w^{(t)}),\, v_M = \mathcal{U}(w^{(t+1)}),\, \mathcal{U}(w) := - \mathcal{L}(w;\, \mathcal{D}_{\textrm{val}})$
			\If {$|v_M - v_0| < \epsilon$}\\
			\hspace{\algorithmicindent}\textbf{return} $SV_k^{(0)} = 0, \, \forall\, k$ (between round truncation)
			\Else
			\For {$\tau = 0,\,1,\,\dots,\,T-1$}
			\For {client $k \in S_t$}
			\State permute $S_t \setminus \{k\}$: $\pi_{\tau}[0] = k,\, \pi_{\tau}[1:M]$
			\State $v_j = v_0$
			\For {$j = 1,\,2,\, \cdots,\, M$}
			\If {$|v_M - v_j| < \epsilon$}
			\State $v_{j+1} = v_j$ (within round truncation)
			\Else
			\State Client Subset $S^{*} = \pi_{\tau}[:j]$
			\State $w^{'} = $ ModelAverage$(n_k,\,w_k^{(t+1)},\,S^{*})$
			\State $v_{j+1} = \mathcal{U}(w^{'}) $
			\EndIf
			\State $SV_{\pi_{\tau}[j]}^{(\tau)} = \frac{\tau-1}{\tau}SV_{\pi_{\tau}[j]}^{(\tau-1)} + \frac{1}{\tau}(v_{j+1} - v_{j})$
			\State $v_{j} = v_{j+1}$
			\EndFor
			\EndFor
			\State \textbf{break} if converged
			\EndFor
			\EndIf\\
			\textbf{return} ${SV_k^{(\tau)}},\, k \in S_t$
		\end{algorithmic}
	\end{algorithm}
	\label{fig:gtgshapleyalgo}
	\vspace{-2\baselineskip}
\end{figure}

\section{Experiments}\label{ss:experiments}
We evaluate our algorithm \textsc{GreedyFed} on several real-world classification tasks against six algorithms: \textsc{FedAvg} \cite{mcmahan2017communication}, \textsc{Power-Of-Choice} \cite{cho2020client}, \textsc{FedProx} \cite{li2020federated}, \textsc{S-FedAvg} \cite{nagalapatti2021game}, \textsc{UCB} \cite{pandey2023goal}  and centralized training at the server (as an upper bound), and demonstrate faster and more stable convergence with higher accuracies under timing constraints from the communication network, especially under heterogeneous settings\footnote{Our code is publicly available at: \url{https://github.com/pringlesinghal/GreedyFed}}

We consider public image datasets MNIST \cite{lecun1998gradient}, FashionMNIST (FMNIST) \cite{xiao2017fashion}, and CIFAR10 \cite{krizhevsky2009learning}. For MNIST and FMNIST, we implement an MLP classifier, while CIFAR10 uses a CNN. For all datasets, we split the original test set into a validation and test set of 5000 images each at the server. 

\textbf{Data Heterogeneity}: We distribute training samples across clients according to the Dirichlet($\alpha$) distribution, where $\alpha$ controls the degree of label distribution skew across clients. The number of data points at each client $n_k = q_k n_{\textrm{train}}$ is distributed according to a power law with $q_k$ sampled from $P(x) = 3x^2, \, x \in (0,1)$ and normalised to sum 1, as done in \cite{cho2020client}. We present results for different values of $\alpha\in \{10^{-4}, 0.1, 100\}$.

\textbf{Timing Constraints}: We impose a fixed budget on the number of communication rounds $T$ for model training. For smaller $T$, model training may be stopped well before convergence if satisfactory performance is achieved. In the real world, such timing constraints may arise for several reasons, like limited communication opportunities with the PS, mobility scenarios, constraints on data usage and power consumption of edge devices, or the need for real-time deployment of models. We compare over $T\in \{150, 250, 350 \}$ for MNIST and FashionMNIST and $T\in \{100, 150, 200\}$ for CIFAR10.

\textbf{Systems Heterogeneity}: FL is characterized by heterogeneity in client-side computing resources. Assuming the presence of another timing constraint for transmitting results to the server within each round, some clients with lower computing or communication resources (stragglers) cannot complete the globally fixed $E$ epochs of training and, thus, transmit partial solutions. We simulate stragglers by selecting $x$ fraction of clients and setting the number of epochs for local training $E_k$ to a uniform random integer in $1$ to $E$, similar to \cite{li2020federated}. We present results for $x\in \{ 0, 0.5, 0.9\}$.

\textbf{Privacy Heterogeneity}: To protect their data, clients can obfuscate transmitted gradients with noise \cite{abadi2016deep}. We simulate varying client privacy requirements with a different noise level for each client. We assign noise level $\sigma_k = (k-1)\frac{\sigma}{N}$ to client $k$ in a random permutation of $[N]$. In each communication round, we add IID Gaussian noise from $\mathcal{N}(0, \sigma_k^2)$ to the updated model parameters of client $k \in S_t$ before transmitting to the PS. We compare over $\sigma\in\{0, 0.05, 0.1\}$.

\textbf{Other Hyperparameters}: We report the mean and standard deviation of test accuracies over five seeds after $T$ communication rounds with $N$ clients and $M$ selections per round. We report $N, M, T$ in each table. We fix $E=5,\, B=5,\, \text{learning rate } \eta=0.01,\, \text{momentum } \gamma=0.5$ across experiments. Data heterogeneity $\alpha = 10^{-4}$ unless specified. We vary \textsc{FedProx} parameter $\mu \in \{0.01, 0.1, 1, 10\}$. 
For \textsc{Power-Of-choice} \cite{cho2020client}, we set an exponential decay rate of $0.9$ on the size of the client query set. For \textsc{GreedyShap}, we consider exponential averaging over $\alpha \in \{0, 0.1, 0.5, 0.9\}$ and standard averaging, and report results for the best parameters. For \textsc{GTG-Shapley} \cite{liu2022gtg} $\epsilon = 10^{-4},T = 50 \times|S|$ for subset $S$, with the default convergence criterion.

\begin{figure}
	\centering
	\includegraphics[width=\linewidth]{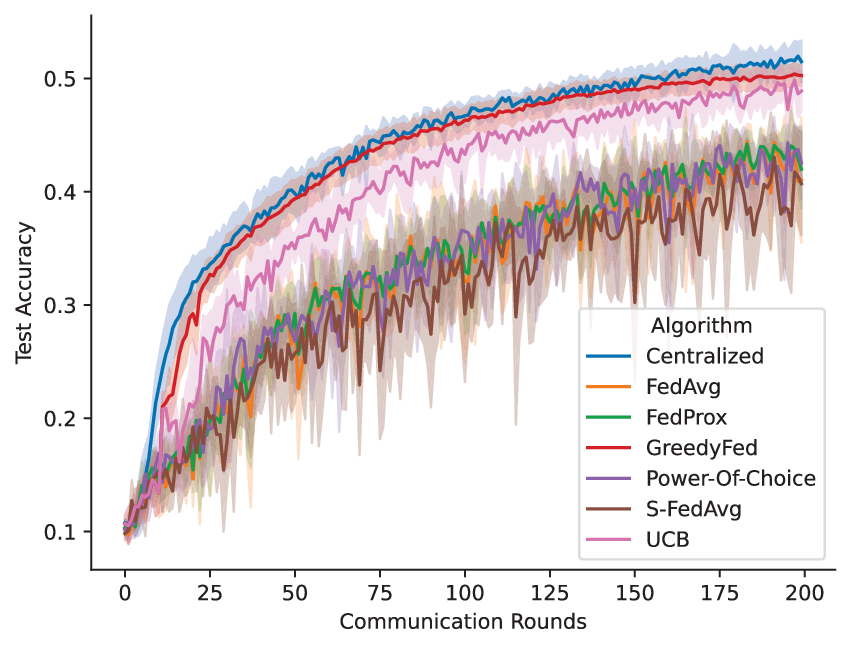}
	\caption{Test Accuracy versus Communication Rounds comparison on CIFAR10, data heterogeneity $\alpha = 10^{-4}$. \textsc{GreedyFed} surpasses all baselines with the fastest convergence and lowest standard deviation, achieving an accuracy close to centralized training.}
	\label{fig:timingplot}
	\vspace{-\baselineskip}
\end{figure}

\section{Analysis}\label{ss:analysis}
\textsc{GreedyFed} consistently achieves higher accuracies and lower standard deviation than all baselines, demonstrating the superiority of our method. All Shapley-Value based methods outperform the other baselines, indicating that this is a promising direction for designing client selection strategies.

\textbf{Data Heterogeneity}: In Table~\ref{tab:datahet}, \textsc{GreedyFed} consistently outperforms baselines in heterogeneous environments ($\alpha = 10^{-4}, 0.1$) and exhibits only a slight performance dip when the data distribution is nearly uniform ($\alpha = 100$).

\textbf{Timing Constraints}: In Table~\ref{tab:timing} \textsc{GreedyFed} achieves faster convergence, with performance saturating for small $T$ (also see Fig.~\ref{fig:timingplot}). For instance, \textsc{GreedyShap} shows only a $4\%$ increase in accuracy on CIFAR10 from $T=150 \text{ to } 350$, while \textsc{FedAvg} shows an $8\%$ increase. 

\textbf{Systems Heterogeneity}: In Table~\ref{tab:systemshet}, \textsc{GreedyFed} shows a smaller drop in performance with a large proportion of stragglers. On FashionMNIST, \textsc{GreedyFed} only experiences a $1\%$ drop in accuracy with $90\%$ stragglers while \textsc{FedAvg} and \textsc{FedProx} experience a $6\%$ drop. More importantly, the standard deviation of other baselines increases drastically in the presence of stragglers, indicating highly unstable training.

\textbf{Privacy Heterogeneity}: In Table~\ref{tab:privacyhet}, \textsc{GreedyFed} shows excellent robustness to noisy clients. On FashionMNIST, \textsc{GreedyFed} only experiences an $8\%$ drop in accuracy for $\sigma = 0.1$ while \textsc{FedAvg} incurs a drastic $20\%$ drop. \textsc{GreedyFed} also shows a much smaller $0.5$ point increase in standard deviation compared to baselines with noise. 

\section{Conclusion}\label{ss:conclusion}
We highlight the benefit of greedy client selection using cumulative Shapley Value in heterogeneous, noisy environments with strict timing constraints on communication. For future work, one can explore how the quality of client selection depends on the distribution of validation data at the server. While our experiments assume similar validation and training distributions, real-world scenarios may involve clients with unique data classes. Previous studies indicate that SV can be unfair to Mavericks (clients with unique data classes), suggesting the need to address distribution mismatches in SV computation and client selection. Future work may also explore providing clients with Shapley Values as feedback encouraging less relevant clients to drop out to reduce communication overhead, similar to \cite{pandey2023goal}.

\begin{table}[t]
	\centering
	\caption{Data Heterogeneity}
	\vspace{-\baselineskip}
	\adjustbox{max width=0.5\textwidth}{
		\begin{tabular}{l|c|c|c|c|c|c|c|c|c}
			\textbf{Dataset} & \textbf{Algorithm} & \textbf{$\alpha = 10^{-4}$} & \textbf{$\alpha = 0.1$} & \textbf{$\alpha = 100$} \\
			\hline
			\multirow{7}{*}{\begin{tabular}[c]{@{}l@{}}\textbf{MNIST} \\ T = 400\\ N = 300\\ M = 3\end{tabular}} 
			& \cellcolor{green!30} GreedyFed & $\mathbf{94.06 \pm 0.19}$ & $88.83 \pm 0.74$ & $93.98 \pm 0.33$ \\
			& UCB & $93.39 \pm 0.72$ & $87.70 \pm 3.19$ & $94.12 \pm 0.28$ \\
			& S-FedAvg & $91.94 \pm 0.95$ & $86.26 \pm 2.10$ & $94.77 \pm 0.29$ \\
			& FedAvg & $92.19 \pm 1.40$ & $\mathbf{88.86 \pm 2.09}$ & $\mathbf{94.79 \pm 0.29}$ \\
			& FedProx & $92.18 \pm 1.39$ & $88.84 \pm 2.12$ & $94.78 \pm 0.32$ \\
			& Power-Of-Choice & $91.58 \pm 1.52$ & $88.27 \pm 2.12$ & $94.72 \pm 0.29$ \\
			& \cellcolor{yellow!50} Centralized & $95.37 \pm 0.26$ & $95.20 \pm 0.24$ & $95.32 \pm 0.22$ \\
			\hline
			\multirow{7}{*}{\begin{tabular}[c]{@{}l@{}}\textbf{FMNIST} \\ T = 400\\ N = 300\\ M = 3\end{tabular}} 
			& \cellcolor{green!30} GreedyFed & $\mathbf{85.18 \pm 0.33}$ & $\mathbf{79.16 \pm 0.62}$ & $84.83 \pm 0.42$ \\
			& UCB & $84.41 \pm 0.68$ & $77.24 \pm 1.24$ & $83.80 \pm 0.73$ \\
			& S-FedAvg & $81.04 \pm 6.09$ & $71.77 \pm 5.87$ & $85.38 \pm 0.38$ \\
			& FedAvg & $82.84 \pm 1.29$ & $72.68 \pm 5.79$ & $85.37 \pm 0.49$ \\
			& FedProx & $82.86 \pm 1.21$ & $73.62 \pm 5.72$ & $85.29 \pm 0.52$ \\
			& Power-Of-Choice & $69.46 \pm 7.34$ & $73.28 \pm 3.45$ & $\mathbf{85.41 \pm 0.22}$ \\
			& \cellcolor{yellow!50} Centralized & $86.16 \pm 0.32$ & $85.75 \pm 0.37$ & $85.95 \pm 0.24$ \\
			\hline
			\multirow{7}{*}{\begin{tabular}[c]{@{}l@{}}\textbf{CIFAR10} \\ T = 200\\ N = 200\\ M = 20\end{tabular}} 
			& \cellcolor{green!30} GreedyFed & $\mathbf{50.26 \pm 0.95}$ & $\mathbf{43.90 \pm 1.04}$ & $50.80 \pm 0.96$ \\
			& UCB & $48.89 \pm 1.84$ & $43.02 \pm 2.75$ & $51.86 \pm 0.62$ \\
			& S-FedAvg & $40.71 \pm 4.54$ & $41.85 \pm 1.59$ & $\mathbf{51.92 \pm 0.58}$ \\
			& FedAvg & $41.04 \pm 5.54$ & $41.67 \pm 3.85$ & $51.74 \pm 0.85$ \\
			& FedProx & $42.00 \pm 3.45$ & $41.66 \pm 3.84$ & $51.74 \pm 0.83$ \\
			& Power-Of-Choice & $42.54 \pm 3.23$ & $42.27 \pm 1.22$ & $51.78 \pm 0.80$ \\
			& \cellcolor{yellow!50} Centralized & $51.48 \pm 1.86$ & $52.36 \pm 1.82$ & $53.22 \pm 1.06$ \\
		\end{tabular}
	}
	\vspace{-\baselineskip}
	\label{tab:datahet}
\end{table}

\begin{table}[t]
	\centering
	\caption{Timing Constraints}
	\vspace{-\baselineskip}
	\adjustbox{max width=0.5\textwidth}{
		\begin{tabular}{l|c|c|c|c|c|c|c|c|c}
			\textbf{Dataset} & \textbf{Algorithm} & \textbf{$T = 150$} & \textbf{$T = 250$} & \textbf{$T = 350$} \\
			\hline
			\multirow{7}{*}{\begin{tabular}[c]{@{}l@{}}\textbf{MNIST} \\ N = 300\\ M = 3\end{tabular}} 
			& \cellcolor{green!30} GreedyFed & $\mathbf{91.12 \pm 0.32}$ & $\mathbf{92.90 \pm 0.28}$ & $\mathbf{93.78 \pm 0.09}$ \\
			& UCB & $90.26 \pm 0.83$ & $90.91 \pm 1.05$ & $92.40 \pm 1.28$ \\
			& S-FedAvg & $87.33 \pm 3.24$ & $91.26 \pm 1.44$ & $89.22 \pm 3.11$ \\
			& FedAvg & $87.56 \pm 2.23$ & $90.20 \pm 3.21$ & $89.50 \pm 0.88$ \\
			& FedProx & $87.57 \pm 2.14$ & $90.22 \pm 3.18$ & $89.53 \pm 0.85$ \\
			& Power-Of-Choice & $85.26 \pm 1.74$ & $89.09 \pm 2.25$ & $91.20 \pm 2.38$ \\
			& \cellcolor{yellow!50} Centralized & $92.77 \pm 0.28$ & $94.27 \pm 0.27$ & $95.04 \pm 0.23$ \\
			\hline
			\multirow{7}{*}{\begin{tabular}[c]{@{}l@{}}\textbf{FMNIST} \\ N = 300\\ M = 3\end{tabular}} 
			& \cellcolor{green!30} GreedyFed & $\mathbf{82.06 \pm 0.51}$ & $\mathbf{84.02 \pm 0.37}$ & $\mathbf{84.75 \pm 0.29}$ \\
			& UCB & $80.53 \pm 2.34$ & $82.53 \pm 1.16$ & $83.04 \pm 1.67$ \\
			& S-FedAvg & $76.25 \pm 4.95$ & $75.04 \pm 6.75$ & $80.23 \pm 5.86$ \\
			& FedAvg & $72.07 \pm 9.28$ & $72.64 \pm 12.22$ & $77.69 \pm 7.64$ \\
			& FedProx & $73.37 \pm 7.10$ & $73.47 \pm 10.06$ & $78.94 \pm 5.66$ \\
			& Power-Of-Choice & $70.13 \pm 5.94$ & $71.74 \pm 9.18$ & $74.93 \pm 12.11$ \\
			& \cellcolor{yellow!50} Centralized & $83.44 \pm 0.38$ & $84.83 \pm 0.55$ & $85.64 \pm 0.41$ \\
			\hline
			&  & \textbf{$T = 100$} & \textbf{$T = 150$} & \textbf{$T = 200$} \\
			\hline
			\multirow{7}{*}{\begin{tabular}[c]{@{}l@{}}\textbf{CIFAR10} \\ N = 200\\ M = 20\end{tabular}} 
			& \cellcolor{green!30} GreedyFed & $\mathbf{46.36 \pm 1.65}$ & $\mathbf{49.38 \pm 1.02}$ & $\mathbf{50.26 \pm 0.95}$ \\
			& UCB & $43.75 \pm 1.44$ & $46.94 \pm 1.74$ & $48.89 \pm 1.84$ \\
			& S-FedAvg & $34.38 \pm 7.00$ & $36.31 \pm 5.78$ & $40.71 \pm 4.54$ \\
			& FedAvg & $33.22 \pm 6.08$ & $40.67 \pm 2.62$ & $41.04 \pm 5.54$ \\
			& FedProx & $34.88 \pm 2.48$ & $40.80 \pm 0.94$ & $42.00 \pm 3.45$ \\
			& Power-Of-Choice & $36.58 \pm 5.22$ & $38.36 \pm 6.18$ & $42.54 \pm 3.23$ \\
			& \cellcolor{yellow!50} Centralized & $46.59 \pm 1.14$ & $49.91 \pm 1.10$ & $51.48 \pm 1.86$ \\
		\end{tabular}
	}
	\label{tab:timing}
	\vspace{-\baselineskip}
\end{table}

\begin{table}[t]
	\centering
	\caption{Systems Heterogeneity}
	\vspace{-\baselineskip}
	\adjustbox{max width=0.5\textwidth}{
		\begin{tabular}{l|c|c|c|c|c|c|c|c|c}
			\textbf{Dataset} & \textbf{Algorithm} & \textbf{$x = 0$} & \textbf{$x = 0.5$} & \textbf{$x = 0.9$} \\
			\hline
			\multirow{7}{*}{\begin{tabular}[c]{@{}l@{}}\textbf{MNIST} \\ T = 400\\ N = 300\\ M = 3\end{tabular}} 
			& \cellcolor{green!30} GreedyFed & $\mathbf{94.06 \pm 0.19}$ & $\mathbf{93.70 \pm 0.33}$ & $\mathbf{92.94 \pm 0.22}$ \\
			& UCB & $93.39 \pm 0.72$ & $92.62 \pm 0.65$ & $91.84 \pm 0.64$ \\
			& S-FedAvg & $91.94 \pm 0.95$ & $91.57 \pm 1.27$ & $92.05 \pm 0.29$ \\
			& FedAvg & $92.19 \pm 1.40$ & $91.03 \pm 1.15$ & $88.44 \pm 5.74$ \\
			& FedProx & $92.18 \pm 1.39$ & $91.05 \pm 1.14$ & $88.45 \pm 5.77$ \\
			& Power-Of-Choice & $91.58 \pm 1.52$ & $89.60 \pm 3.43$ & $90.24 \pm 2.19$ \\
			& \cellcolor{yellow!50} Centralized & $95.37 \pm 0.26$ & $95.37 \pm 0.26$ & $95.37 \pm 0.26$ \\
			\hline
			\multirow{7}{*}{\begin{tabular}[c]{@{}l@{}}\textbf{FMNIST} \\ T = 400\\ N = 300\\ M = 3\end{tabular}} & \cellcolor{green!30} GreedyFed & $\mathbf{85.18 \pm 0.33}$ & $\mathbf{84.66 \pm 0.55}$ & $\mathbf{84.19 \pm 0.19}$ \\
			& UCB & $84.41 \pm 0.68$ & $83.58 \pm 1.15$ & $81.86 \pm 2.56$ \\
			& S-FedAvg & $81.04 \pm 6.09$ & $82.23 \pm 2.11$ & $74.66 \pm 10.17$ \\
			& FedAvg & $82.84 \pm 1.29$ & $78.93 \pm 5.20$ & $76.20 \pm 6.53$ \\
			& FedProx & $82.86 \pm 1.21$ & $79.35 \pm 4.61$ & $76.31 \pm 6.01$ \\
			& Power-Of-Choice & $69.46 \pm 7.34$ & $81.49 \pm 3.12$ & $79.03 \pm 4.63$ \\
			& \cellcolor{yellow!50} Centralized & $86.16 \pm 0.32$ & $86.16 \pm 0.32$ & $86.16 \pm 0.32$ \\
			\hline
			\multirow{7}{*}{\begin{tabular}[c]{@{}l@{}}\textbf{CIFAR10} \\ T = 200\\ N = 200\\ M = 20\end{tabular}} 
			& \cellcolor{green!30} GreedyFed & $\mathbf{50.26 \pm 0.95}$ & $\mathbf{48.74 \pm 1.34}$ & $\mathbf{47.58 \pm 0.77}$ \\
			& UCB & $48.89 \pm 1.84$ & $47.98 \pm 1.71$ & $46.51 \pm 0.87$ \\
			& S-FedAvg & $40.71 \pm 4.54$ & $38.55 \pm 5.93$ & $34.88 \pm 3.54$ \\
			& FedAvg & $41.04 \pm 5.54$ & $43.16 \pm 2.57$ & $37.23 \pm 4.47$ \\
			& FedProx & $42.00 \pm 3.45$ & $43.26 \pm 2.54$ & $37.21 \pm 4.35$ \\
			& Power-Of-Choice & $42.54 \pm 3.23$ & $36.92 \pm 8.51$ & $37.39 \pm 3.41$ \\
			& \cellcolor{yellow!50} Centralized & $51.48 \pm 1.86$ & $51.52 \pm 1.85$ & $51.52 \pm 1.94$ \\
		\end{tabular}
	}
	\vspace{-\baselineskip}
	\label{tab:systemshet}
\end{table}

\begin{table}[t]
	\centering
	\caption{Privacy Heterogeneity}
	\vspace{-\baselineskip}
	\adjustbox{max width=0.5\textwidth}{
		\begin{tabular}{l|c|c|c|c|c|c|c|c|c}
			\textbf{Dataset} & \textbf{Algorithm} & \textbf{$\sigma = 0$} & \textbf{$\sigma = 0.05$} & \textbf{$\sigma = 0.1$} \\
			\hline
			\multirow{7}{*}{\begin{tabular}[c]{@{}l@{}}\textbf{MNIST} \\ T = 400\\ N = 300\\ M = 3\end{tabular}} 
			& \cellcolor{green!30} GreedyFed & $\mathbf{94.06 \pm 0.19}$ & $\mathbf{90.79 \pm 0.45}$ & $\mathbf{87.33 \pm 0.68}$ \\
			& UCB & $93.39 \pm 0.72$ & $87.23 \pm 1.82$ & $75.36 \pm 4.67$ \\
			& S-FedAvg & $91.94 \pm 0.95$ & $79.17 \pm 5.88$ & $62.78 \pm 4.37$ \\
			& FedAvg & $92.19 \pm 1.40$ & $83.98 \pm 2.45$ & $63.63 \pm 6.34$ \\
			& FedProx & $92.18 \pm 1.39$ & $83.95 \pm 2.42$ & $63.33 \pm 6.31$ \\
			& Power-Of-Choice & $91.58 \pm 1.52$ & $83.87 \pm 1.85$ & $58.22 \pm 4.55$ \\
			& \cellcolor{yellow!50} Centralized & $95.37 \pm 0.26$ & $95.37 \pm 0.26$ & $95.37 \pm 0.26$ \\
			\hline
			\multirow{7}{*}{\begin{tabular}[c]{@{}l@{}}\textbf{FMNIST} \\ T = 400\\ N = 300\\ M = 3\end{tabular}} 
			& \cellcolor{green!30} GreedyFed & $\mathbf{85.18 \pm 0.33}$ & $\mathbf{81.23 \pm 0.55}$ & $\mathbf{77.17 \pm 0.73}$ \\
			& UCB & $84.41 \pm 0.68$ & $76.11 \pm 2.28$ & $67.26 \pm 7.29$ \\
			& S-FedAvg & $81.04 \pm 6.09$ & $75.45 \pm 3.72$ & $57.67 \pm 7.49$ \\
			& FedAvg & $82.84 \pm 1.29$ & $74.77 \pm 2.12$ & $62.18 \pm 4.39$ \\
			& FedProx & $82.86 \pm 1.21$ & $74.81 \pm 2.06$ & $62.79 \pm 4.71$ \\
			& Power-Of-Choice & $69.46 \pm 7.34$ & $69.47 \pm 5.08$ & $56.08 \pm 8.06$ \\
			& \cellcolor{yellow!50} Centralized & $86.16 \pm 0.32$ & $86.16 \pm 0.32$ & $86.16 \pm 0.32$ \\
			\hline
			\multirow{7}{*}{\begin{tabular}[c]{@{}l@{}}\textbf{CIFAR10} \\ T = 200\\ N = 200\\ M = 20\end{tabular}} 
			& \cellcolor{green!30} GreedyFed & $\mathbf{50.26 \pm 0.95}$ & $\mathbf{48.44 \pm 1.06}$ & $\mathbf{47.64 \pm 1.02}$ \\
			& UCB & $48.89 \pm 1.84$ & $47.86 \pm 1.87$ & $44.76 \pm 2.19$ \\
			& S-FedAvg & $40.71 \pm 4.54$ & $39.41 \pm 4.07$ & $31.89 \pm 12.33$ \\
			& FedAvg & $41.04 \pm 5.54$ & $39.74 \pm 6.30$ & $36.34 \pm 5.53$ \\
			& FedProx & $42.00 \pm 3.45$ & $39.66 \pm 6.27$ & $36.16 \pm 5.52$ \\
			& Power-Of-Choice & $42.54 \pm 3.23$ & $38.55 \pm 2.82$ & $36.19 \pm 1.87$ \\
			& \cellcolor{yellow!50} Centralized & $51.48 \pm 1.86$ & $51.44 \pm 1.83$ & $51.46 \pm 1.93$ \\
		\end{tabular}
	}
	\label{tab:privacyhet}
	\vspace{-\baselineskip}
\end{table}

\bibliographystyle{ieeetr}
\bibliography{ref}
\end{document}